\documentclass[letterpaper, 10 pt, journal, twoside]{IEEEtran}
\IEEEoverridecommandlockouts

\usepackage[english]{babel}
\usepackage[utf8]{inputenc}
\usepackage[T1]{fontenc}
\usepackage{comment}

\usepackage[dvipsnames]{xcolor}
\usepackage{graphicx}
\usepackage{subcaption}
\usepackage{multirow}
\usepackage{multicol}

\usepackage{amsmath}
\usepackage{amssymb}
\usepackage{booktabs} %

\usepackage{footmisc}
\usepackage{placeins} %
\usepackage{url}

\usepackage{pifont} %

\usepackage{xcolor,colortbl}
\definecolor{Gray}{gray}{0.88}
\definecolor{DarkGray}{gray}{0.78}
\newcolumntype{h}{>{\columncolor{DarkGray}}c}
\newcolumntype{g}{>{\columncolor{Gray}}c}

\usepackage{hyperref}
\PassOptionsToPackage{unicode}{hyperref}
\PassOptionsToPackage{naturalnames}{hyperref}

\usepackage[absolute]{textpos}

\newif\ifdraft
\draftfalse
\newcommand{\parag}[1]{\paragraph{#1}}

\newcommand{\bP}{\mathbf{P}}

\newcommand{\mytitle}{Perspective Aware Road Obstacle Detection}
\newcommand{\myauthor}{Krzysztof Lis, Sina Honari, Pascal Fua, and Mathieu Salzmann}
\newcommand{\mydoi}{10.1109/LRA.2023.3245410}

\hypersetup{
	pdftitle={\mytitle},
	pdfauthor={\myauthor},
	pdfkeywords={Computer Vision for Transportation, Data Sets for Robotic Vision, Deep Learning for Visual Perception, Object Detection, Segmentation and Categorization},
	pdfinfo={
		DOI={\mydoi},
	}
}

\begin{document}

\title{\mytitle}

\author{\myauthor%
\thanks{Manuscript received: September 13, 2022; Revised December 21, 2022; Accepted February 9, 2023. %
This paper was recommended for publication by  by Associate Editor I. Gilitschenski and Editor C. Cadena Lerma upon evaluation of the reviewers' comments.
The work was supported in part by the International Chair Drive for All - MINES ParisTech - Peugeot-Citroën - Safran - Valeo. 
}
\thanks{All authors are with Computer Vision Laboratory, EPFL, Lausanne, Switzerland. 
{\footnotesize (krzysztof.lis@epfl.ch, lis.krzysztof@protonmail.com; sina.honari@gmail.com; pascal.fua@epfl.ch; mathieu.salzmann@epfl.ch)}.
}
\thanks{Digital Object Identifier (DOI): \mydoi}
}

\markboth{IEEE Robotics and Automation Letters. Preprint Version. Accepted February, 2023}{Lis \MakeLowercase{\textit{et al.}}: \mytitle} 

\maketitle

\IEEEpeerreviewmaketitle

\begin{abstract}
While road obstacle detection techniques have become increasingly effective, they typically ignore the fact that, in practice, the apparent size of the obstacles decreases as their distance to the vehicle increases. 
In this paper, we account for this  by computing a scale map encoding the apparent size of a hypothetical object at every image location. We then leverage this perspective map to 
(i) generate training data by injecting onto the road synthetic objects whose size corresponds to the perspective foreshortening; 
and (ii) incorporate perspective information in the decoding part of the detection network to guide the obstacle detector. 
Our results on standard benchmarks show that, together, these two strategies significantly boost the obstacle detection performance, allowing our approach to consistently outperform state-of-the-art methods in terms of instance-level obstacle detection. 
\end{abstract}


\begin{textblock*}{20cm}(0.8cm,26.8cm) %
\footnotesize\noindent
© 2023 IEEE.  Personal use of this material is permitted.  Permission from IEEE must be obtained for all other uses, in any current or future media, including reprinting/republishing this material for advertising or promotional purposes, creating new collective works, for resale or redistribution to servers or lists, or reuse of any copyrighted component of this work in other works.
\end{textblock*}

\section{Introduction}\label{sec:intro}

\IEEEPARstart{V}{ision-based} driving assistance is now commercially available~\cite{LaneAssist} and enables vehicles to plan a path within the predicted drivable space while avoiding other traffic. However, unusual and unexpected obstacles lying on the road remain a potential danger. 
Since not every vehicle has stereo cameras or a LiDAR sensor to detect them in 3D, much effort has recently been made to achieve detection in a monocular fashion via learning-based strategies. Such road obstacle detection can also be beneficial for robots in novel environments.
Given that such objects are non-exclusive, obtaining exhaustive datasets of real images annotated with such obstacles for training purposes is impractical. Hence, many state-of-the-art deep learning approaches~\cite{Lis19,DiBlase21,Lis20,Chan21a} rely on synthetically-generated training data, e.g., by cutting out objects and inserting them into individual frames of the Cityscapes dataset.

\definecolor{RedCircle}{rgb}{0.3, 0., 0.}
\definecolor{GreenCircle}{rgb}{0., 0.3, 0.}

\begin{figure} 
\centering
\begin{tabular}{cc}
\vspace{-3mm}  & \multirow{2}{*}{\includegraphics[width=0.572\linewidth]{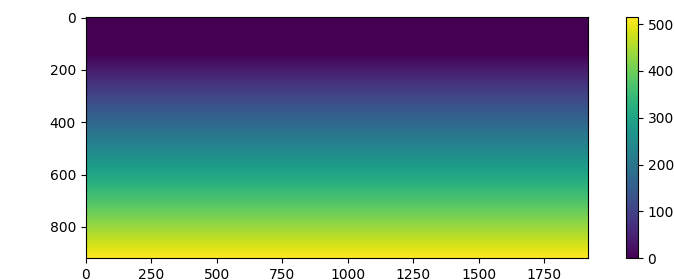}} \\
\includegraphics[width=0.43\linewidth]{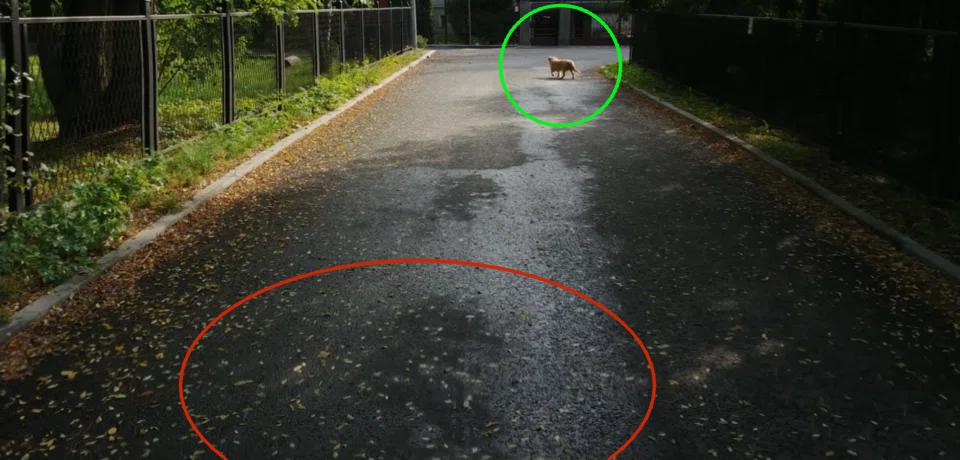} & {} \\
(a)&(b) \\
\includegraphics[width=0.43\linewidth]{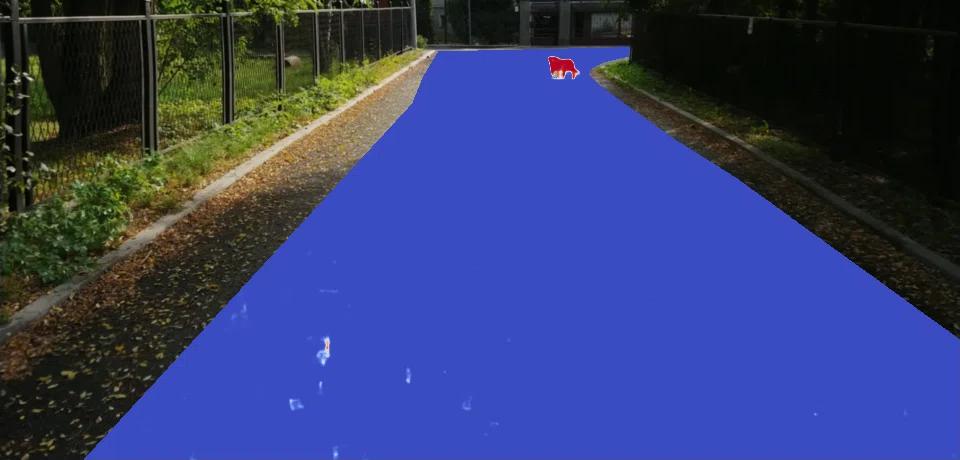}&
\includegraphics[width=0.43\linewidth]{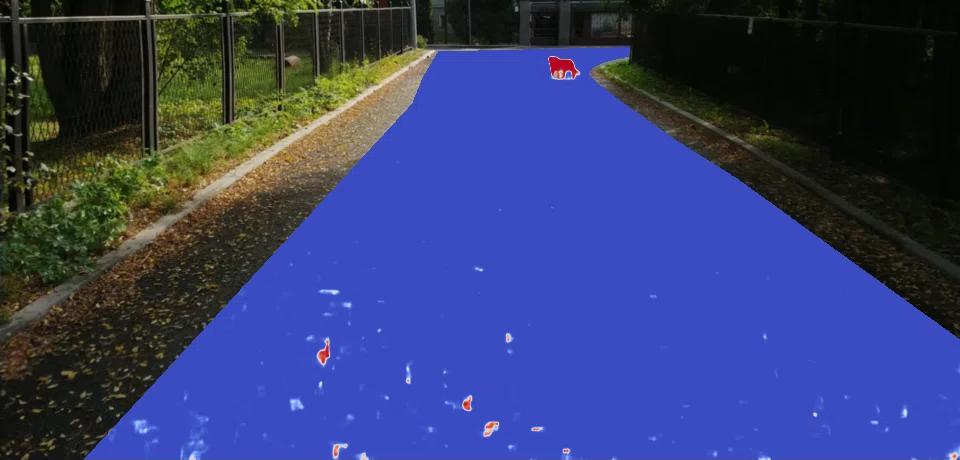}\\
(c)&(d)
\end{tabular}
\vspace{-3mm}
\caption{\small{\bf Far and relevant vs close and irrelevant.} (a) Original image. The {\color{green}green circle} denotes a real obstacle far away, and the {\color{red} red circle} indicates nearby but harmless leaves. (b) The perspective map indicates, at each pixel, the size in pixels of a hypothetical meter-wide object at that location.
(c) Our approach uses the perspective map to distinguish relevant objects from irrelevant ones.
It correctly flags in red the pixels of the real obstacle while ignoring the leaves. (d) Without the {\it perspective aware training set}, a network with a similar architecture flags them all.}
\vspace{-3mm}
\label{fig:teaser}
\end{figure}

However, these methods fail to leverage, both while generating training data and performing the actual detection, the predictable perspective foreshortening in images captured by vehicles' front-facing cameras. It is a standard practice~\cite{Lis20,Bevandic19,Vojir21} to insert objects of arbitrary sizes at any image location in the training data and to detect objects at multiple-scales irrespective of where they appear in the image. This does not exploit the well-known fact that more distant objects tend to be smaller and that, given a calibrated camera, the relationship between real and projected sizes is known.

In this work, we show that leveraging the perspective information substantially increases performance. To this end, as shown in Fig.~\ref{fig:teaser}, we compute a scale map, whose pixel values denote the apparent size in pixels of a hypothetical meter-wide object placed at that point on the road. We then exploit this information in two complementary ways:
\begin{itemize}

	\vspace{-1mm}
	\item{\bf Perspective-Aware Synthetic Object Injection.} Instead of uniformly injecting synthetic objects into road scenes to synthesize training data, as in~\cite{Lis20,Bevandic19,Vojir21}, we use the perspective map to appropriately set the projected size of the objects we insert.
	
	\vspace{-0mm}
	\item{\bf Perspective-Aware Architecture.}  We feed the perspective map at multiple levels of a feature pyramid network, enabling it to learn the realistic relationship between distance and size embodied in our training set and in real road scenes.

\end{itemize}
The bottom portion of Fig.~\ref{fig:teaser} illustrates the benefits of our approach. It not only detects small far-away obstacles but also avoids false alarms arising from small irregularities near the car, such as the leaves here, because their size at this image location does not match that of real threats to the vehicle. Our results show that these strategies together contribute to significantly improving the accuracy of road obstacle detection, particularly in terms of instance-level detection, which is critical for a self-driving car that need to identify all potential hazards on the road.

We evaluate our approach on the \textit{Segment Me If You Can}~\cite{Chan21b} benchmark's obstacle track and the {\it Lost\&Found}~\cite{Pinggera16} test subset. We demonstrate that it significantly outperforms state-of-the-art techniques that use architectures similar to ours, but without explicit perspective handling.
The implementation of our method is available at \url{https://github.com/cvlab-epfl/perspective-aware-obstacles}.

\section{Related Work}
\label{sec:related}

A complete overview of state-of-the-art road anomaly detection methods can be found in~\cite{Chan21b}. In short, many of the most effective monocular methods, as ours, generate synthetic training data to palliate for the lack of a sufficiently diverse annotated road obstacle dataset. We therefore focus on these methods, and then discuss other attempts at exploiting perspective information for diverse tasks.

\subsection{Synthetic Training Data for Obstacle Detection}

There is an intractable variety of unexpected objects that can pose a collision threat on roads. To handle this diversity, most existing obstacle detection methods rely on creating synthetic data for training purposes. It is often created from background traffic frames, often from Cityscapes~\cite{Cordts16}, into which synthetic obstacles are inserted.

In~\cite{Lis19}, the synthetic anomalies are generated by altering the semantic class of existing object instances and synthesizing an image from those altered labels. 
In~\cite{DiBlase21}, this is complemented by adding the Cityscapes {\it void} regions as obstacles. 
However, many of the objects exploited by these techniques are located above or away from the road, and the resulting training data only yields limited performance for small on-road obstacles. Our results show that we outperform these methods. 

In~\cite{Vojir21}, synthetic obstacles are obtained by cropping random polygons within the background frame and copying their content onto the road, or filling them with a random color. Other methods~\cite{Lis20,Bevandic19,Grcic22} inject object instances extracted from various image datasets. 
While this can be done effectively, it remains suboptimal because the objects are placed at random locations, without accounting for their size or for the scene geometry. This is what we address here by explicitly exploiting perspective information, and we demonstrate that it yields a substantial performance boost.

\subsection{Exploiting Perspective Information}

Earlier works~\cite{Hadsell07, Hadsell09} propose a lightweight sliding-window classifier of drivable space
using a pyramid of input patches whose dimension depends on their distance from the horizon.
These patches are then rescaled according to their distance to the camera, 
ensuring that the similar obstacles have similar pixels sizes when presented to the classifier, regardless of the effects of perspective in the original image.
This application of perspective information to overcome scale variance is effective,
but it can not be easily combined with standard CNNs which operate on the whole image rather than individually rescaled patches.

For any perspective camera, distortion depends on image  position. A popular approach to enabling a deep network to account for this in its predictions is to provide it with pixel coordinates as input. In~\cite{Liu18e, Ulyanov18, Lyu18}, this is achieved by treating normalized pixel coordinates as two additional channels.
In~\cite{Choi20} the pixel coordinates are used to compute an attention map, to exploit the fact that the class distribution correlates with the image height, for example the {\it sky} class is predominantly at the top of the image.
Another way to implicitly account for perspective effects is to introduce extra network branches that process the image at different scales and fuse the results~\cite{Li17e,Huynh21}. However, this strategy, as those relying on pixel coordinates, does not explicitly leverage the perspective information available when working with a calibrated camera, as is typically the case in self-driving.

None of obstacle-detection algorithms explicitly accounts for the relationship between projected object size and distance. This can be done by creating {\it scale maps} that encode the expected size in the world of an image pixel depending on its position. 
Scale maps have been used for obstacle and anomaly detection~\cite{Bai19b,Prakash19}. In~\cite{Bai19b}, 
the scale information is used to crop and resize image regions before passing them to a vehicle detection network, which then gets to view the cars at an approximately constant scale. This requires running the detector multiple times on the crops. By contrast, our method processes the whole image at once, and the model learns how to leverage the perspective information to adjust the features. In~\cite{Prakash19}, the scale maps are used to rectify the road surfaces, and obstacles are then detected in the rectified views. Unlike these methods, we exploit perspective maps as input to our network, instead of using them for image pre-processing.  
This prevents the creation of visual artifacts caused by image warping, which yields higher accuracy, as we will show in experiments.

Scale maps have also been extensively investigated for crowd counting purposes~\cite{Chan08,Shi19a,Zhang15c,Yang20a,Liu19a,Liu19b}. In~\cite{Chan08,Shi19a}, the models predict perspective information based on observed body and head size. In~\cite{Zhang15c}, an unsupervised meta-learning method is deployed to learn perspective maps, which are then used to warp the input images so that they depict a uniform scale as in \cite{Yang20a}. In~\cite{Liu19a}, a scale map serves as an extra channel alongside the RGB image and is passed through the backbone feature extractor, whereas in~\cite{Liu19b} an additional branch is added to the  backbone to process the single-channel scale map and to concatenate the resulting features afterwards. In short, perspective information is used during feature computation. In this paper, we follow a different track and incorporate the scale map at different levels of a feature pyramid network. Our experiments show this to be more effective. Furthermore, we argue and demonstrate that, for anomaly detection, incorporating perspective information into the network is not enough; one must also exploit it when synthesizing training data.

\subsection{Fusing RGB and Depth for Road Obstacle Detection}
When depth information is available, from stereo camera disparity or RGB-D sensors,
it can be fused with the RGB appearance to improve obstacle detection. For example,
\cite{Ramos17} combines semantic segmentation with stereo-based detections;
MergeNet~\cite{Gupta18a} extracts complementary features from RGB-D; RFNet~\cite{Sun20}'s two-stream backbone extracts RGB and depth features and uses them to output joint segmentation of known classes and unusual obstacles.
Depth (or disparity) contains important geometric cues about the obstacles, which protrude from the road plane, and the above-mentioned methods exploit these cues to detect the obstacles.
By contrast, we do not use stereo images and the associated precise scene geometry; our perspective map is generated using a flat-road assumption and contains no information about the obstacles.
Our architecture uses perspective as context for analyzing obstacle appearance, by taking it as an extra feature channel without any processing.

\section{Approach}\label{sec:method}

Our approach relies on a {\it perspective map} that captures scale change of objects on the road plane.
Therefore, we also refer to it as  a \textit{scale map}.
In this section, we first describe its construction and then how we use it both to control training data synthesis and as an input to our detection network.

\subsection{Computing the Perspective Map}
\label{sec:perspective}

\begin{figure} 
\centering
\includegraphics[width=0.52\linewidth]{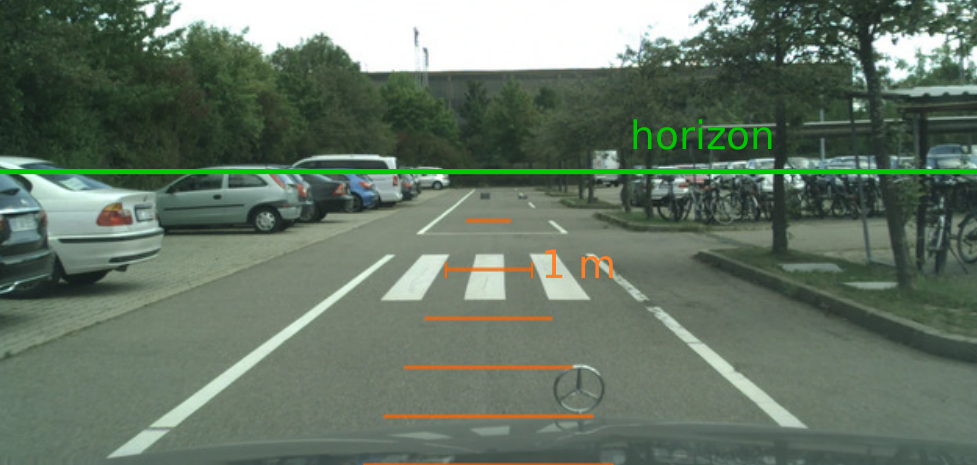}
\includegraphics[width=0.36\linewidth]{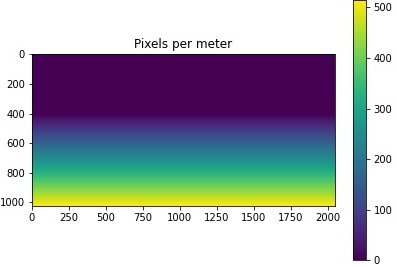}
\caption{\small{\bf Perspective map.}{\it Left:} A 1-meter length overlaid at different image heights on an image from the {\it Lost\&Found} dataset.
{\it Right:} The corresponding perspective map. 
}
\label{fig:perspective_map}
\end{figure}

A perspective map is a scalar field whose value at a given pixel denotes the width in pixels of a hypothetical meter-wide object placed at that point on the road. 
Fig.~\ref{fig:perspective_map} depicts one. We compute it from the camera calibration parameters, which are known in a self-driving setup because a vehicle's camera can be calibrated during its production. The camera parameters are $f$, the camera's focal length in pixels, $H$, its elevation above the ground in meters, and $\theta$, its pitch angle.

\begin{figure} 
\centering
\includegraphics[width=0.7\linewidth]{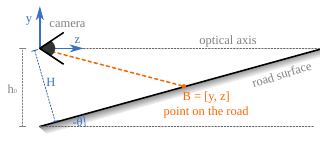}
\caption{\small{\bf Building the perspective map.} 
Geometry of a front-facing camera viewing a planar road surface. 
Here, the $y$ coordinate of the orange point is negative because it is below the optical axis. 
}
\label{fig:camera_geometry}
\vspace{-2mm}
\end{figure}

We assume the road to be planar, which, locally, is a good approximation in the majority of real driving scenarios.
Let us consider a 3D road point $B=[x, y, z]$ in camera coordinates, which projects onto $[u, v]$ in the image space, as shown in Fig.~\ref{fig:camera_geometry}. For simplicity, we denote by $[0, 0]$ the principal point that lies at the center of the image, which is also known in a calibrated camera. Assuming the road to be planar, 
the pinhole camera model dictates that 
\begin{equation}
	v = f \frac{y}{z} \; . 
	\label{eq:proj}
\end{equation}
As $B$ lies on the road plane, which is inclined by an angle $\theta$ w.r.t. the camera's optical axis, we can write
\begin{equation}
	y = z \tan(\theta) - h_0 \; , 
	\label{eq:y}
\end{equation}
where $h_0 = \frac{H}{\cos{\theta}}$, with $H$ being the perpendicular distance from the camera to the road. Solving for $z$ by replacing $y$ in Eq.~\ref{eq:proj} by  its definition in Eq.~\ref{eq:y} yields 
\begin{equation}
	z([u,v]) =  h_0 \frac{f}{ f\tan(\theta) - v}  \;,
\end{equation}
where we indicate that $z$ depends on the pixel location $[u,v]$ by $z([u,v])$. The visible scale is inversely proportional to the $z$ coordinate in the camera frame, and so the scale value $P([u,v])$ in the perspective map  $\bP$ is equal to
\begin{equation}
	P([u,v]) = f\frac{1}{z([u,v])} = \frac{\cos(\theta)}{H} (f \tan(\theta) - v)\;.
\end{equation}
Note that this requires the pitch angle $\theta$ of the camera optical axis with respect to the road surface to be known. When the car is stable on its four wheels, it only depends on how the camera is mounted, which is known. 
If the pitch changes while driving, an online camera calibration module, e.g., one that relies on vanishing points~\cite{Chaudhury14}, could be used to update the value of  $\theta$.
We also assume that various distortions have been corrected so that a pinhole camera model applies. 

\subsection{Perspective-Aware Synthetic Object Injection}
\label{sec:Synthetic-Inject}

Collecting a training database of all items that could potentially be left on the road and pose a collision threat is impractical. Effective obstacle detection can thus only be achieved via handling previously unseen objects. To this end, existing methods~\cite{Lis20,Vojir21,Bevandic19} generate synthetic training frames by injecting objects into the road scenes. However, they use random object sizes and locations. Instead, we leverage the perspective map so that the inserted object sizes are consistent with their locations on the road plane.

\begin{figure} 
\centering
\includegraphics[width=0.49\linewidth]{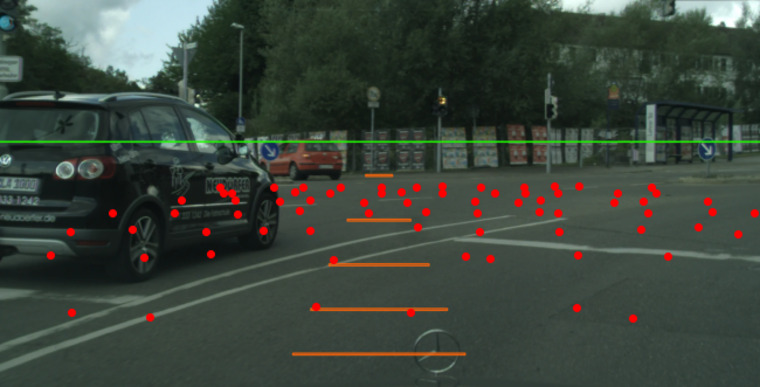}
\includegraphics[width=0.49\linewidth]{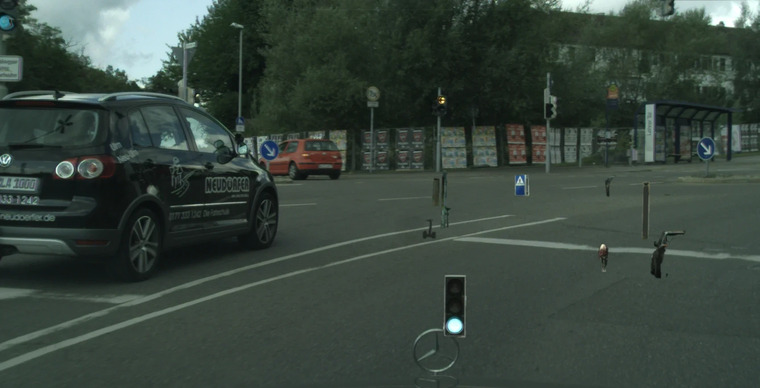}
\caption{\small{\it Left:} Anchor points distributed along the road surface. We place obstacles at a random subset of anchor points.
{\it Right:} Frame with injected obstacles.
}
\vspace{-4mm}
\label{fig:synth_grid}
\end{figure}

\parag{Placement} We generate a rectangular grid on the road plane, with grid lines being 3.5 meters apart in the direction along the road and 1 meter apart in the width direction. Once this grid is projected onto the image, each grid intersection yields an anchor point offset from the grid point by a random vector whose coordinates are drawn from a zero-centered normal distribution with standard deviation of 0.5 meter. The anchor points are shown in Fig.~\ref{fig:synth_grid}. We then place obstacles at a random subset of these anchor points.

\begin{figure} 
\centering
\begin{tabular}{cc}
\hspace{-5mm}
\includegraphics[width=0.49\linewidth]{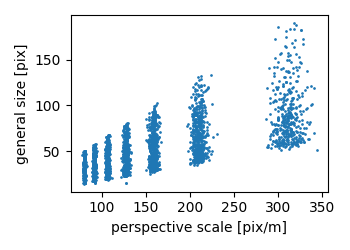}&
\includegraphics[width=0.49\linewidth]{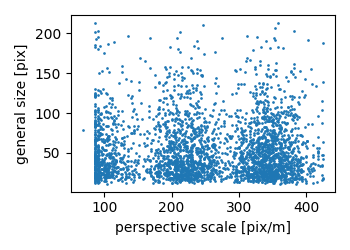} \\
Perspective Aware & \; \; \; \; \; Uniform~\cite{Lis20}
\end{tabular}
\vspace{-0.5mm}
\caption{\small 
Distribution of injected object sizes.
\textit{Left:} Our \textbf{Perspective Aware Strategy} 
selects object sizes based on the perspective map 
to ensure that objects look smaller when they are further away.
There are clusters because we inject objects at discrete grid points on the road-plane projected to the image. 
\textit{Right:} The \textbf{Uniform Strategy} chooses random objects from the whole instance database.
}
\label{fig:synth_distribution}
\vspace{-2pt}
\end{figure}

\parag{Size}
We extract object instances -- vehicles, pedestrians, traffic signs and lights -- from the Cityscapes dataset~\cite{Cordts16}. 
These yield image cut-outs of diverse shapes, ranging from thin poles to wide vehicles.
We take an object's {\it overall pixel size} $\text{pix}_{\text{obj}}$ to be the average of three values: the square root of its pixel area, and its bounding box width and height.
We aim to generate synthetic objects
within a range of physical sizes $[\text{ph}_\text{min}, \text{ph}_\text{max}]$ in meters.
To simulate the corresponding visible pixel size of an object seen at an image point $[u, v]$, we multiply the physical range by the scale map value at $[u, v]$: 
$\text{pix}_{(\text{min},\text{max})}([u, v]) = \text{ph}_{(\text{min},\text{max})} P([u, v])$.
Then, we randomly select an object from the training set whose size satisfies
$\text{pix}_{\text{min}}([u, v])  \leq \text{pix}_\text{obj} \leq \text{pix}_\text{max}([u, v])$ and paste it at $[u, v]$.
Since we use known classes, such as humans or cars, to later detect unknown classes, such as bottles or tires, we ignore the original size of the known items, and instead choose the object size as a hyper-parameter using the validation-set.
Fig.~\ref{fig:synth_grid}-right depicts the resulting object insertion. Note that this does \emph{not} involve scaling the original cut-out objects. Instead, we simply select objects of the appropriate size, thus avoiding scaling artifacts. In Fig.~\ref{fig:synth_distribution}, we visualize the resulting relationship between object size and perspective map for both our perspective-aware approach and a uniform injection one. We will compare these two approaches quantitatively. %

\subsection{Perspective-Aware Architecture}
\label{sec:arch}
To distinguish obstacle pixels from the road surface ones, we rely on a U-Net type network architecture, which we train using negative binary cross-entropy loss of pixel classification between the model's prediction and the ground truth segmentation map. The input image is first processed by a ResNeXt101~\cite{Xie17b} feature extractor, pre-trained on ImageNet~\cite{Deng09} and frozen at training time. We extract four levels of features with increasing receptive fields.

\begin{figure} 
\centering
\includegraphics[width=0.95\linewidth]{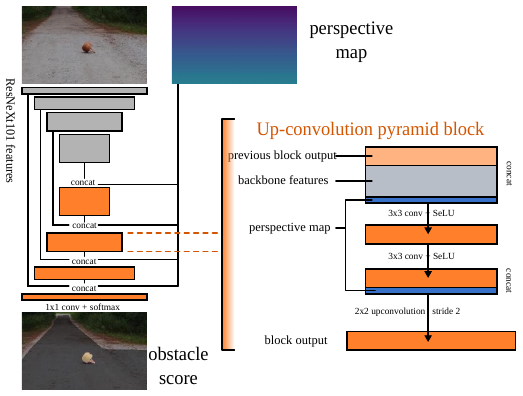}
\caption{\small Perspective-aware architecture. 
The perspective map is injected into the decoding blocks at different resolutions. In each block it is appended twice; first to the backbone features, and second to the intermediate activations preceding the transpose convolution for upsampling.}
\label{fig:arch_small}
\vspace{-10pt}
\end{figure}

In each block of our up-convolution pyramid, we concatenate the perspective map to the backbone features, and we use it again before the transposed convolution, as depicted in Fig.~\ref{fig:arch_small}. With such an insertion, the scale information presented to each level of the pyramid can then influence the interpretation of the backbone features at different receptive fields and hence locally adjust the effective receptive field of the detector. In practice, this allows the network to distinguish between distant obstacles and small but harmless irregularities, such as wet patches, leaves, or tile edges, which ought to be ignored. As evidenced below, our perspective-aware architecture shows its full advantage when used together with the perspective-aware object injection.

In our experiments, the perspective map is scaled by $\frac{1}{400}$ before being passed to the network, following the normalization applied in~\cite{Liu19b}, to bring it to an approximate value range between 0 and 1, which improves convergence. 

\section{Experiments}\label{sec:experiments}
In this section, we present the datasets and metrics, and compare our method with the state-of-the-art ones.

\begin{table*}
\centering
\setlength{\heavyrulewidth}{2pt}
\scalebox{0.72}{
\begin{tabular}{l|c|hcc|c||hcc|c}
\toprule
& & \multicolumn{4}{c||}{Road Obstacles 21} & \multicolumn{4}{c}{Lost\&Found - Test No Known} \\
\specialrule{\lightrulewidth}{0pt}{0pt}
& requires & \multicolumn{3}{c}{Component-level} & \multicolumn{1}{|c||}{Pixel-level} & \multicolumn{3}{c}{Component-level} & \multicolumn{1}{|c}{Pixel-level} \\
Method & OOD & $\overline{F_1}\uparrow$ & $\overline{\mathrm{sIoU}}$ $\uparrow$ & $\overline{\mathrm{PPV}}$ $\uparrow$ & AuPRC $\uparrow$ & $\overline{F_1}\uparrow$ & $\overline{\mathrm{sIoU}}$ $\uparrow$ & $\overline{\mathrm{PPV}}$ $\uparrow$ & AuPRC $\uparrow$ \\
\specialrule{\lightrulewidth}{0pt}{0pt}
Ours & No & \textbf{67.1 $\pm$ 1.7} & \textbf{65.2 $\pm$ 0.6} & 60.2 $\pm$ 2.7 & 75.2 $\pm$ 0.1 & \textbf{68.6 $\pm$ 0.4} & \textbf{49.8 $\pm$ 0.6} & \textbf{87.6 $\pm$ 1.2} & \textbf{87.4 $\pm$ 0.4} \\
\specialrule{\lightrulewidth}{0pt}{0pt}
DenseHybrid \cite{Grcic22} & Yes & 50.7 & 45.7 & 50.1 & \textbf{87.1} & 52.3 & 46.9 & 52.1 & 78.7 \\ %
Maximized Entropy \cite{Chan21a} & Yes & 48.5 & 47.9 & \textbf{62.6} & 85.1 & 49.9 & 45.9 & 63.1 & 77.9 \\
SynBoost \cite{DiBlase21} & Yes & 37.6 & 44.3 & 41.8 & 71.3 & 48.7 & 36.8 & 72.3 & 81.7  \\
Road Inpainting \cite{Lis20} & No & 36.0 & 57.6 & 39.5 & 54.1 & 52.3 & 49.2 & 60.7 & 82.9 \\
JSRNet \cite{Vojir21} & No & 11.0 & 18.6 & 24.5 & 28.1 & 36.0 & 34.3 & 45.9 & 74.2  \\ %
ODIN \cite{Liang18b} & No & 9.4 & 21.6 & 18.5 & 22.1 & 34.5 & 39.8 & 49.3 & 52.9  \\
Image Resynthesis \cite{Lis19} & No & 8.4 & 16.6 & 20.5 & 37.7 & 19.2 & 27.2 & 30.7 & 57.1  \\
Maximum Softmax \cite{Hendrycks17b} & No & 6.3 & 19.7 & 15.9 & 15.7 & 10.3 & 14.2 & 62.2 & 30.1 \\
Void Classifier \cite{Blum19} & Yes & 5.4 & 6.3 & 20.3 & 10.4 & 1.9 & 1.8 & 35.1 & 4.8  \\
Mahalanobis \cite{Lee18a} & No & 4.7 & 13.5 & 21.8 & 20.9 & 22.1 & 33.8 & 31.7 & 55.0  \\
Embedding Density  \cite{Blum19} & No & 2.3 & 35.6 & 2.9 & 0.8 & 27.5 & 37.8 & 35.2 & 61.7  \\
Ensemble  \cite{Lakshminarayanan17}  & No & 1.3 & 8.6 & 4.7 & 1.1 & 2.7 & 6.7 & 7.6 & 2.9  \\
MC Dropout \cite{Mukhoti18} & No & 1.0 & 5.5 & 5.8 & 4.9 & 13.0 & 17.4 & 34.7 & 36.8 \\
\bottomrule
\end{tabular}
}
\caption{\small 
Obstacle detection scores on RoadObstacle21 and  Lost\&Found datasets. 
Both component-level and pixel-level metrics are reported on each dataset. 
The primary metric is average component detection $\overline{F_1}$ score. 
\textit{Requires OOD} column indicates if a model is using out of distribution data by training on additional datasets. Other methods train only on Cityscapes dataset.
}
\label{tab:metrics_baselines_short}
\vspace{-5pt}
\end{table*}

\begin{table*}[hbt!]
\centering
\setlength{\heavyrulewidth}{2pt}
\scalebox{0.72}{
\begin{tabular}{l|c|hc||hc}
\toprule
& & \multicolumn{2}{c||}{Road Obstacles 21} & \multicolumn{2}{c}{Lost\&Found - Test No Known} \\
\specialrule{\lightrulewidth}{0pt}{0pt}
Architecture & Object Injection & \multicolumn{1}{c}{$\overline{F_1}\uparrow$} & \multicolumn{1}{c||}{AuPRC $\uparrow$} & \multicolumn{1}{c}{$\overline{F_1}\uparrow$} & \multicolumn{1}{c}{AuPRC $\uparrow$} \\
\specialrule{\lightrulewidth}{0pt}{0pt}
1 Ours (perspective-aware) & Ours (perspective-aware) & \textbf{67.1 $\pm$ 1.7} & 75.2 $\pm$ 0.1 & \textbf{68.6 $\pm$ 0.4} & \textbf{87.4 $\pm$ 0.4} \\
2 No perspective channel & Ours (perspective-aware) & 52.5 $\pm$ 6.7 & 70.6 $\pm$ 1.0 & 63.5 $\pm$ 3.9 & 86.0 $\pm$ 0.8 \\
3 P-map backbone branch~\cite{Liu19b} & Ours (perspective-aware) & 65.0 $\pm$ 2.2 & 74.6 $\pm$ 1.2 & 63.5 $\pm$ 2.4 & 85.5 $\pm$ 0.7 \\
4 P-map along RGB~\cite{Liu19a} & Ours (perspective-aware) & 58.2 $\pm$ 4.1 & 64.6 $\pm$ 2.8 & 66.2 $\pm$ 3.5 & 73.9 $\pm$ 8.3 \\
5 XY channels & Ours (perspective-aware) & 54.8 $\pm$ 4.2 & 71.1 $\pm$ 2.3 & 63.8 $\pm$ 0.8 & 86.1 $\pm$ 0.1 \\
6 Image warping~\cite{Prakash19,Yang20a} & Ours (perspective-aware) & 45.2 $\pm$ 0.5 & 65.5 $\pm$ 0.7 & 20.3 $\pm$ 0.6 & 43.5 $\pm$ 1.3 \\ \hline

7 Ours (perspective-aware) & Uniform~\cite{Lis20} & 56.1 $\pm$ 1.7 & \textbf{77.1 $\pm$ 0.9} & 52.4 $\pm$ 2.6 & 82.1 $\pm$ 3.3 \\
8 No perspective channel & Uniform~\cite{Lis20} & 43.7 $\pm$ 6.6 & 73.3 $\pm$ 1.7 & 48.5 $\pm$ 6.4 & 74.5 $\pm$ 10.2 \\
9 P-map backbone branch~\cite{Liu19b} & Uniform~\cite{Lis20} & 53.3 $\pm$ 4.3 & 76.8 $\pm$ 0.5 & 55.2 $\pm$ 1.5 & 84.3 $\pm$ 0.2 \\
10 P-map along RGB~\cite{Liu19a} & Uniform~\cite{Lis20} & 50.8 $\pm$ 1.6 & 69.0 $\pm$ 2.4 & 50.6 $\pm$ 5.8 & 79.7 $\pm$ 1.5 \\
11 XY channels & Uniform~\cite{Lis20} & 51.9 $\pm$ 1.4 & 75.7 $\pm$ 0.5 & 53.7 $\pm$ 2.1 & 78.7 $\pm$ 1.7 \\

\bottomrule
\end{tabular}

\quad
\begin{tabular}{rl|cc|cc}
	\toprule
	\multicolumn{2}{c|}{Object size [m]} & \multicolumn{2}{c|}{Road Obstacles 21 - Validation} & \multicolumn{2}{c}{Lost\&Found - Train} \\
	\specialrule{\lightrulewidth}{0pt}{0pt}
	$\text{ph}_\text{min}$ & $\text{ph}_\text{max}$ & \multicolumn{1}{c}{$\overline{F_1}\uparrow$} & AuPRC $\uparrow$ & $\overline{F_1}\uparrow$ & AuPRC $\uparrow$ \\
	\specialrule{\lightrulewidth}{0pt}{0pt}
	\specialrule{\lightrulewidth}{0pt}{0pt}
	0.1 &- 0.3 & 59.3 $\pm$ 3.4 & 80.2 $\pm$ 1.6 & \textbf{66.1 $\pm$ 1.1} & 77.5 $\pm$ 0.5 \\
	0.25 &- 0.55 & 65.1 $\pm$ 3.0 & 95.7 $\pm$ 0.7 & 62.5 $\pm$ 0.3 & \textbf{89.4 $\pm$ 0.8} \\
	0.5 &- 0.9 & \textbf{65.6 $\pm$ 1.4} & \textbf{96.6 $\pm$ 0.4} & 57.5 $\pm$ 0.1 & 87.6 $\pm$ 0.4 \\
	0.75 &- 1.25 & 49.5 $\pm$ 1.7 & 94.2 $\pm$ 0.1 & 50.3 $\pm$ 2.4 & 86.8 $\pm$ 1.4 \\
	all & sizes~\cite{Lis20} & 56.1 $\pm$ 1.7 & 77.1 $\pm$ 0.9 & 52.4 $\pm$ 2.6 & 82.1 $\pm$ 3.3 \\
	\bottomrule
\end{tabular}
}
\caption{\small 
Ablation study.
Left) We compare different variants of utilizing the perspective map and show their impact while  using either uniform or our perspective-aware object injection. Right) Effect of the size of the injected training objects.
}
\label{tab:metrics_ablation_short}
\vspace{-10pt}
\end{table*}

\subsection{Datasets}\label{sec:exp_datasets}
We train our network on Cityscapes. During training, we sample patches of 768 $\times$ 384 pixels, with random horizontal flipping. We also perform noise augmentation so that the network generalizes to road surfaces rougher than those found in Cityscapes. 
We follow the evaluation protocol of the {\it Segment Me If You Can}\cite{Chan21b} obstacle detection benchmark and test our network using the following two datasets.

\parag{Lost \& Found - Test No Known}
Lost \& Found~\cite{Pinggera16} is an established obstacle dataset captured by placing objects on the road and taking images from an approaching vehicle. The {\it No Known} variant excludes objects present in the Cityscapes training set, such as pedestrians or bicycles, to focus on the methods' ability to generalize to previously unseen obstacles. It contains 1043 frames with 1709 occurrences of 7 unique lost cargo items placed in 12 parking lot and street scenes. The camera calibration parameters required to compute the perspective map are part of the dataset. 

\parag{RoadObstacle21}
RoadObstacles21 is the obstacle track of the recent {\it Segment Me} benchmark~\cite{Chan21b}. Like Lost \& Found, it contains photos of obstacles placed on roads, but it expands the number of unique objects and the diversity of the scenes to include more road textures and weather conditions. It comprises 327 frames containing 388 occurrences of 31 unique objects placed in 8 scenes. There are no camera calibration parameters. We therefore estimated them as follows. We assume $f = 2265 \mbox{pix}$, that is, the same focal length as in the Cityscapes training set, and $H = 1.5 \mbox{m}$ because the dataset was captured using handheld cameras.  
We estimate the camera's pitch angle by approximating the horizon level - the image-space position of the road plane's vanishing line.
In the considered datasets, the camera has no side-to-side roll, so we assume the line to be horizontal.
The sides of the road and not regular enough to fit lines to them, but with
the images depicting forward views along the roads, the horizon is slightly above the end of the visible road. 
Hence, we first segmented the road using the semantic segmentation PSP network~\cite{Zhao17b}, and then took the approximate horizon level to be 16 pixels above its uppermost edge. 
Given $v_\textnormal{horiz}$, the number of pixels between the image midpoint and the horizon level, the pitch angle is retrieved as
\[
	\theta = \tan^{-1}(\frac{v_\textnormal{horiz}}{f}) \; .
\]
Such an estimate is obviously very rough, but it is sufficient to inform our model of the scale changes on the road, as we will empirically show when comparing to other variants of our model that do not leverage perspective information.

\subsection{Metrics}\label{sec:exp_metrics}

The {\it Segment Me} benchmark~\cite{Chan21b} measures the methods' performance at both pixel and component levels. The pixel classification task involves distinguishing pixels belonging to obstacles from those of the road surface. It is primarily evaluated with the area under the precision-recall curve (AuPRC). However, pixel metrics give more importance to nearby and big obstacles than to distant or small ones, because of the image area they occupy. For the purpose of driving safety, it thus is more relevant to reason in terms of obstacle instances and their detection regardless of distance and image size. This is addressed by component-level metrics, such as $\overline{F_1}$, $\overline{\mbox{sIoU}}$ and $\overline{\mbox{PPV}}$, proposed in \cite{Chan21b}.

\subsection{Quantitative Evaluation}\label{sec:exp_baselines}

In Table~\ref{tab:metrics_baselines_short}, we compare our approach to the state-of-the-art methods featured in the {\it Segment Me} benchmark. The {\it requires OOD (out of distribution)} column indicates whether the method was trained using additional data beyond the commonly-used Cityscapes training set, for example by using objects from COCO~\cite{Lin14a} as obstacles. Our method outperforms the baselines in terms of instance metrics, and only performs worse than \cite{Chan21a} and \mbox{\cite{Grcic22}} on two metrics, which leverages OOD, thereby demonstrating the good generalization of our approach without resorting to extra training data.

\subsection{Ablation study}
\label{sec:exp_ablation}

\subsubsection{Impact of Perspective}
\label{sec:ablation_perspective}

In Table~\ref{tab:metrics_ablation_short}-left, we report the results of an ablation study in which we altered either our architecture to use the perspective map in different ways or the perspective-aware synthetic object insertion strategy.

In particular, we consider the following variants:

\begin{itemize}
	\item{\it Ours (perspective-aware)}: Our full architecture using the perspective map as described in Section \ref{sec:arch}.
	\item{\it No perspective channel} : This variant omits the perspective map from our complete architecture.
	\item{\it P-map backbone branch} : We provide the perspective map as input to the network but process it in a feature extractor separately from the RGB feature extractor, as in the architecture of \cite{Liu19b}.
	\item{\it P-map along RGB}: We provide the perspective map along with RGB as input to the network, following \cite{Liu19a}. In this variant, we 
	unfreeze the backbone feature extractor weights during training to let the network train on the perspective inputs.
	\item{\it XY channels} : We provide the image coordinates as two additional channels instead of the perspective map, applying the idea from~\cite{Liu18e,Choi20}.
	\item{\it Image warping}: We follow the idea of \cite{Prakash19,Yang20a} and use the perspective map to transform the image into a top-down view of the road.
\end{itemize}

\begin{figure*}
\centering
\rotatebox{90}{\parbox[c][0.07\linewidth][c]{60pt}{\centering \tiny \hspace{0.0cm} Image}}
\includegraphics[width=0.300\linewidth]{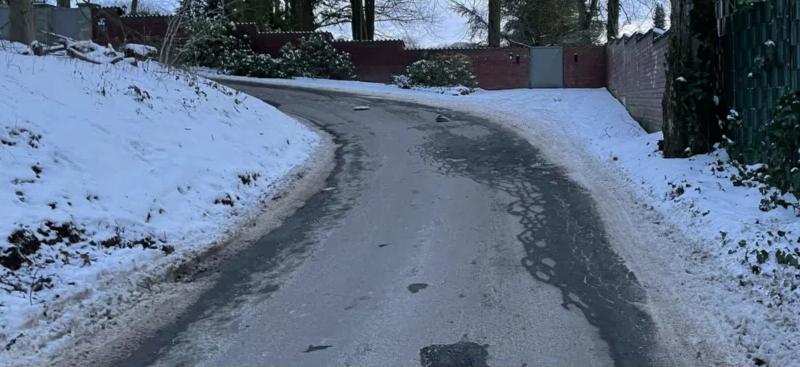}
\includegraphics[width=0.300\linewidth]{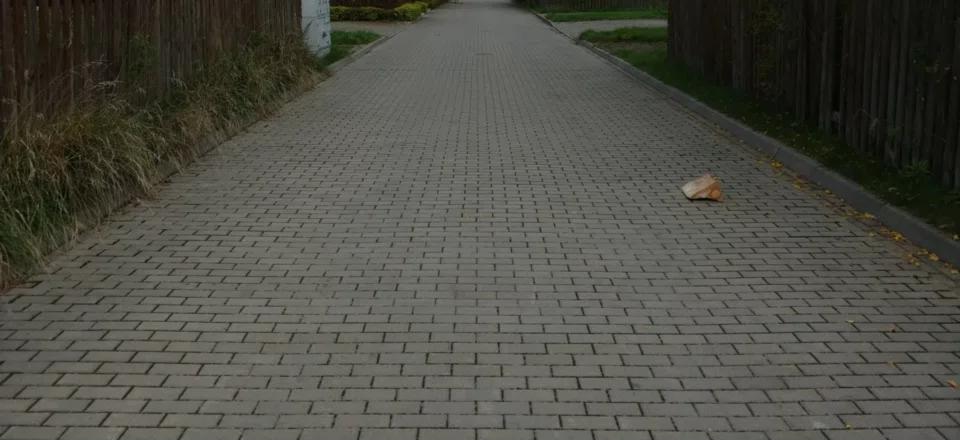}
\includegraphics[width=0.300\linewidth]{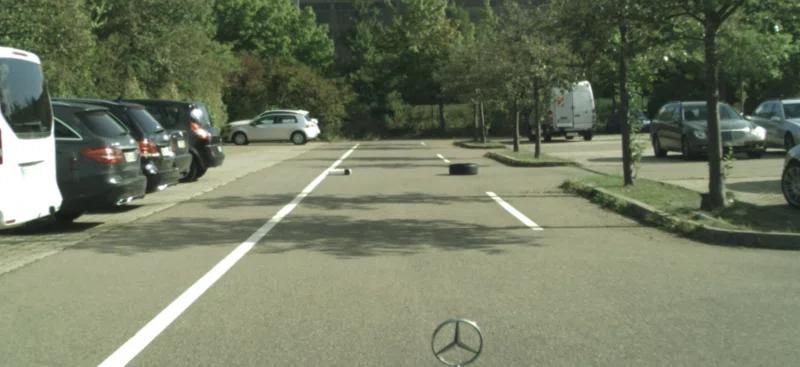}

\vspace{1.5mm}
\rotatebox{90}{\parbox[c][0.07\linewidth][c]{60pt}{\centering \tiny \hspace{0.0cm} Ours}}
\includegraphics[width=0.300\linewidth]{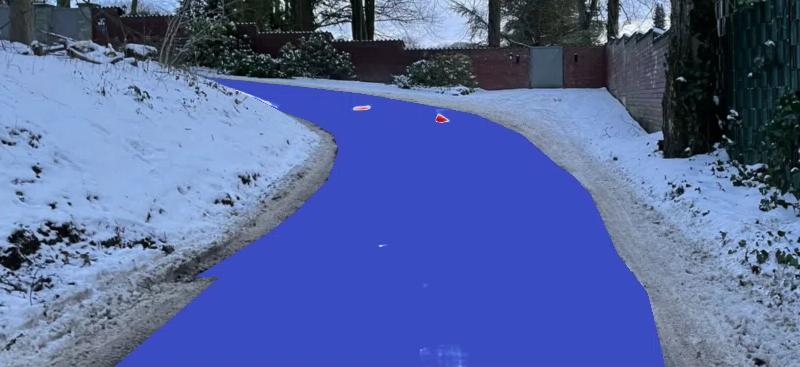}
\includegraphics[width=0.300\linewidth]{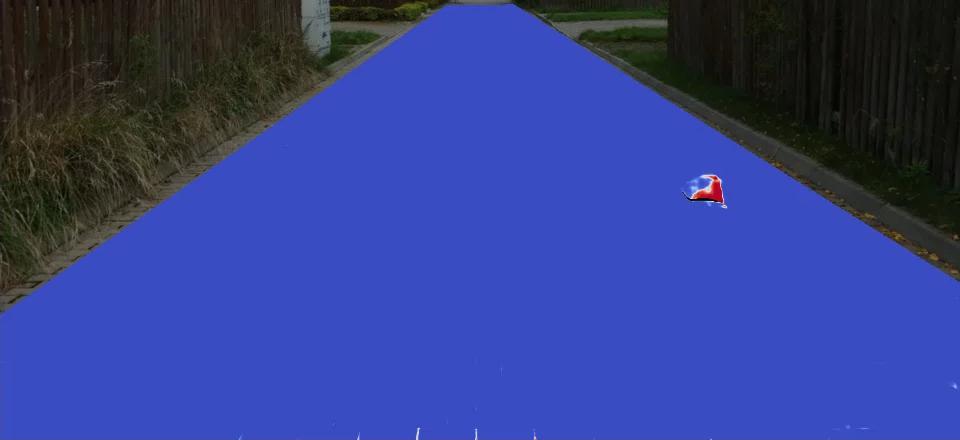}
\includegraphics[width=0.300\linewidth]{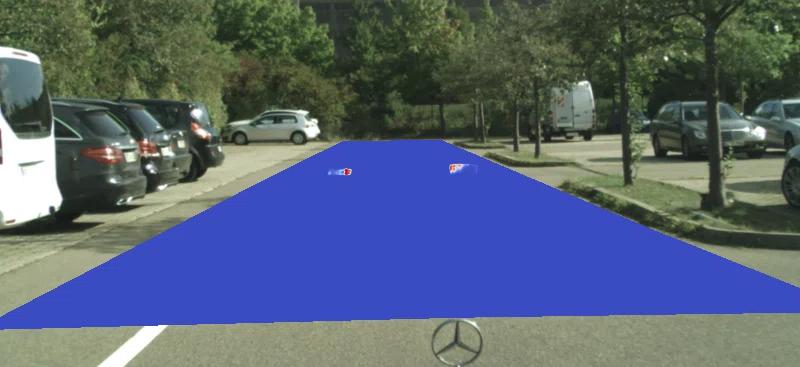}

\vspace{1.5mm}
\rotatebox{90}{\parbox[c][0.07\linewidth][c]{60pt}{\centering \tiny \hspace{0.0cm} No perspective \\ \hspace{0.0cm} aware synth~\cite{Lis20}}}
\includegraphics[width=0.300\linewidth]{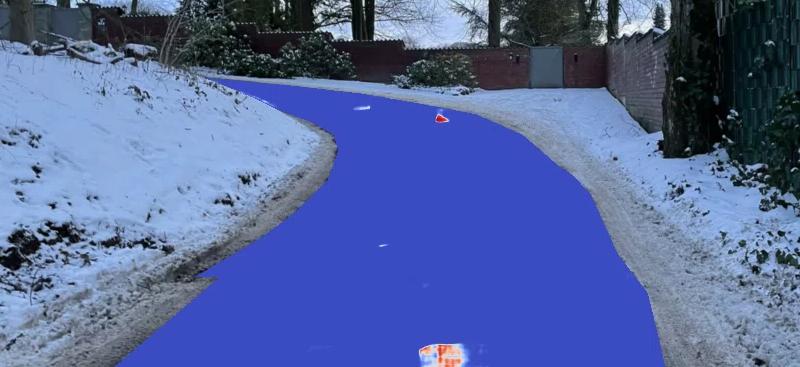}
\includegraphics[width=0.300\linewidth]{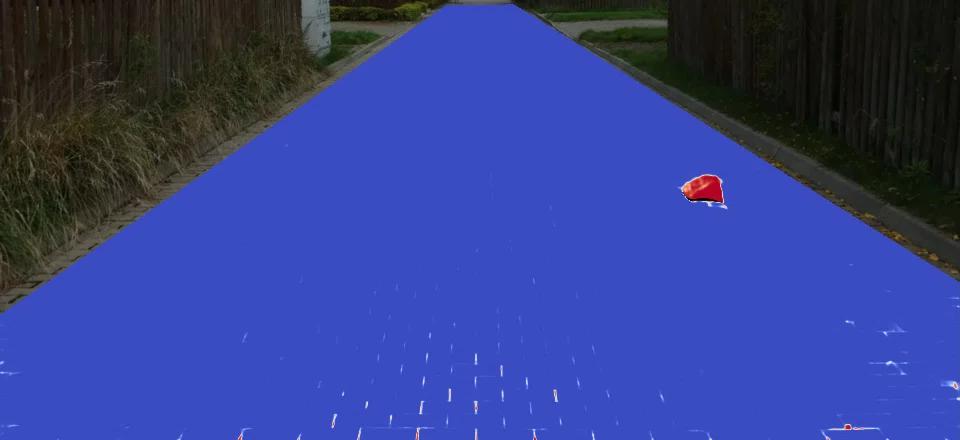}
\includegraphics[width=0.300\linewidth]{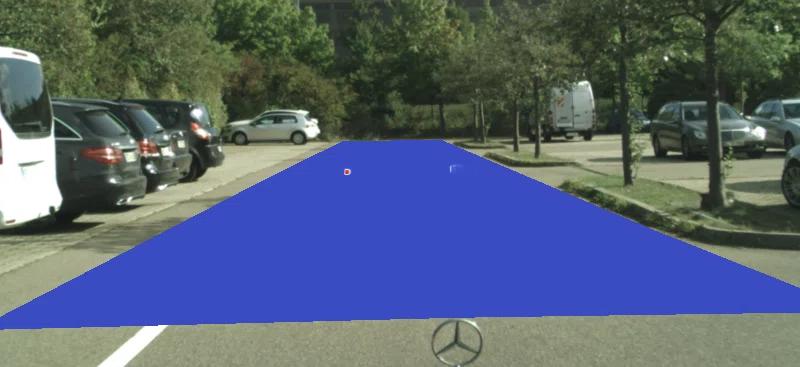}

\vspace{1.5mm}
\rotatebox{90}{	\parbox[c][0.07\linewidth][c]{60pt}{\centering \tiny \hspace{0.0cm} Max entropy~\cite{Chan21a}}} 
\includegraphics[width=0.300\linewidth]{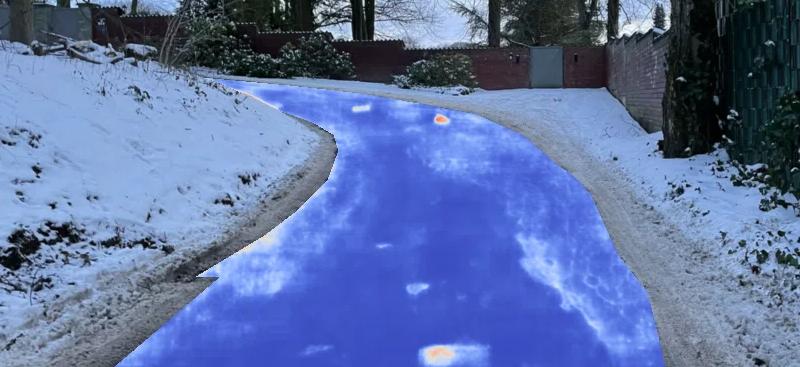}
\includegraphics[width=0.300\linewidth]{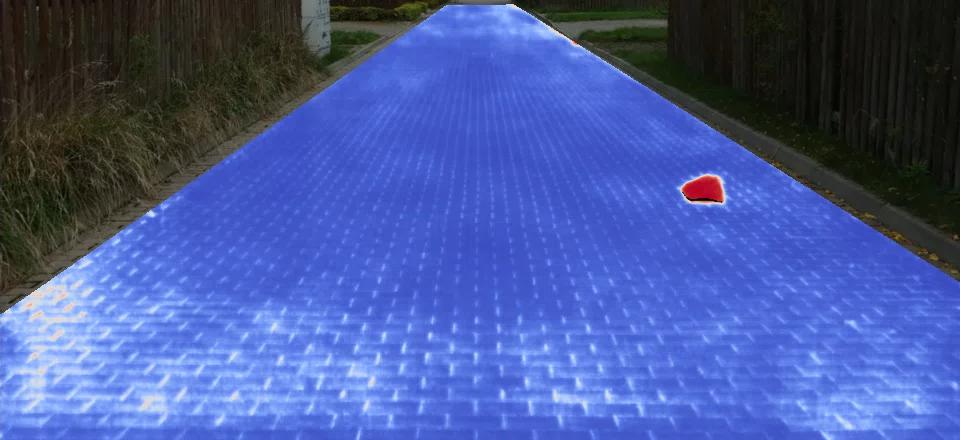}
\includegraphics[width=0.300\linewidth]{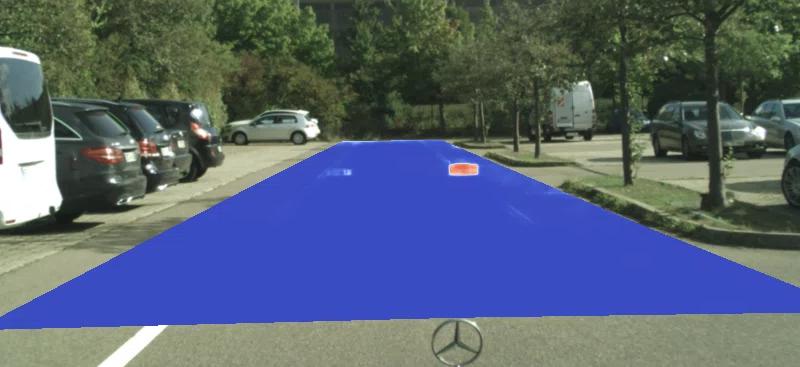}
\caption{\small{\it Left, center: } Perspective information guides our detector to ignore nearby small irregularities on the road surface, while the variants without perspective map and perspective-aware object insertion exhibit false positives in that area.
The nearby false-positives and distant obstacles are of similar pixel sizes, so the perspective map allows differentiating between them.
{\it Right:} Our method finds both obstacle instances despite imperfect segmentation. While Max entropy~\cite{Chan21a} achieves a better pixel-classification score by perfectly segmenting the bigger object,
it misses the smaller object.
}
\label{fig:qualitative}
\end{figure*}

For synthetic object insertion, we consider two variants; one with perspective-aware object injection as described in Section \ref{sec:Synthetic-Inject}, and another with uniform object insertion that injects objects uniformly from the full object pool, without restricting it to the objects whose size are inversely proportional to the location where they will be placed.
In this variant, the obstacles are placed uniformly in image space rather than on a grid in the road plane.
This strategy is identical to~\cite{Lis20}.%

In each setup we provide the mean and standard deviations over three training runs.
As observed in Table~\ref{tab:metrics_ablation_short}-left, using our perspective-aware objection injection together with our perspective-aware network architecture yields the best performance (row 1), and dropping each one of them reduces the accuracy (rows 2, 7, 8). It is worth noting that using the {\it P-map backbone branch} architecture with our perspective-aware object insertion (row 3) also yields close, yet still inferior results, which indicates that, while there can be other ways of using the perspective map in the architecture, the perspective-aware object insertion plays a key role. %
The {\it No perspective channel} variant shows a noticeable drop (row 7), indicating that the network benefits from exploiting the perspective maps.
While using the XY image coordinates (row 5) yields reasonable results in {\it Lost\&Found}, whose camera angle matches that of the training set, its performance drops when faced with the different camera setup of {\it Road Obstacles}.
The {\it P-map along RGB} and {\it Image warping} variants also show a significant drop (rows 4, 6), which indicates that the way the perspective map is used plays a role in the network accuracy. In particular, the performance of {\it Image warping}, where the network operates on the warped image, is much lower. %

The results show that training with a uniform injection strategy (rows 7-11) yields a much lower $\overline{F_1}$ score than with our perspective-aware approach (rows 1-5). 
However, the pixel classification AuPRC values of the uniform strategy tend to be higher.
We observe that the networks trained with the uniform object insertion technique often better segment the large objects (predicting more pixels on the object),
but miss small objects and introduce small false positive instances.
By contrast, the higher $\overline{F_1}$ scores obtained with our perspective-aware injection approach evidence the resulting networks' reliability in detecting obstacles more accurately and avoiding false-positives, which is more critical for self-driving cars.

We show qualitative examples in Fig.~\ref{fig:qualitative}. The first two columns evidence how the perspective map helps the model to distinguish distant obstacles from nearby harmless details, which can span similar pixel sizes in the image. In the rightmost column, we show a difficult case where Max-entropy~\cite{Chan21a} obtains higher AuPRC but lower $\overline{F_1}$ than us. While such models can segment more pixels of the found objects, they entirely miss some of the objects on the road. 
Overall, our network detects the obstacles more reliably, and learns to ignore small irregular regions. %

\begin{figure*}
\centering
\includegraphics[width=0.38\linewidth]{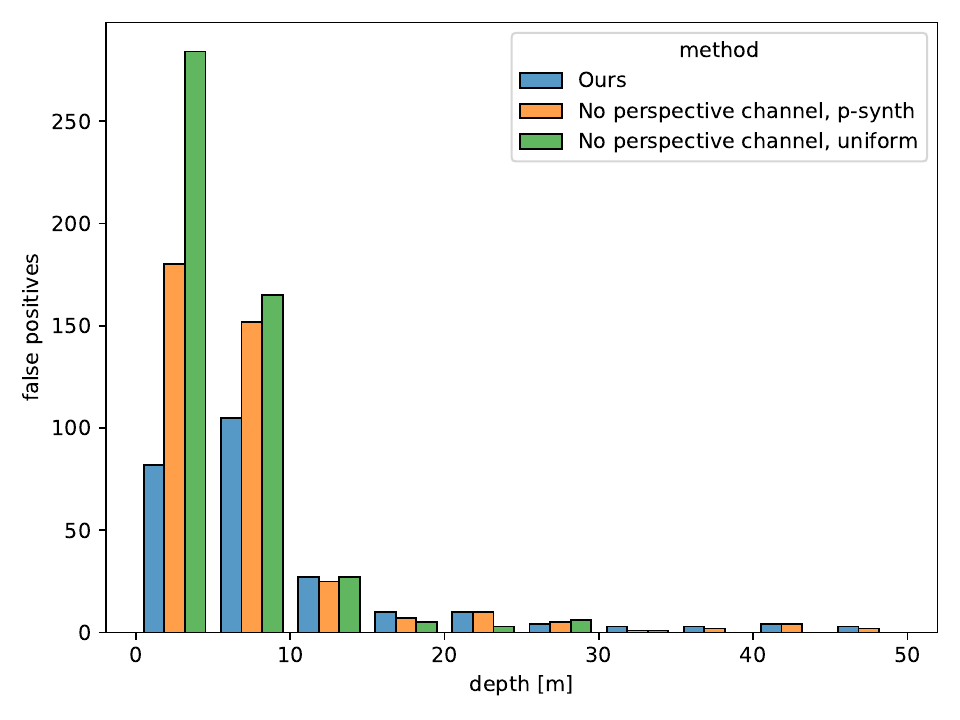}
\includegraphics[width=0.38\linewidth]{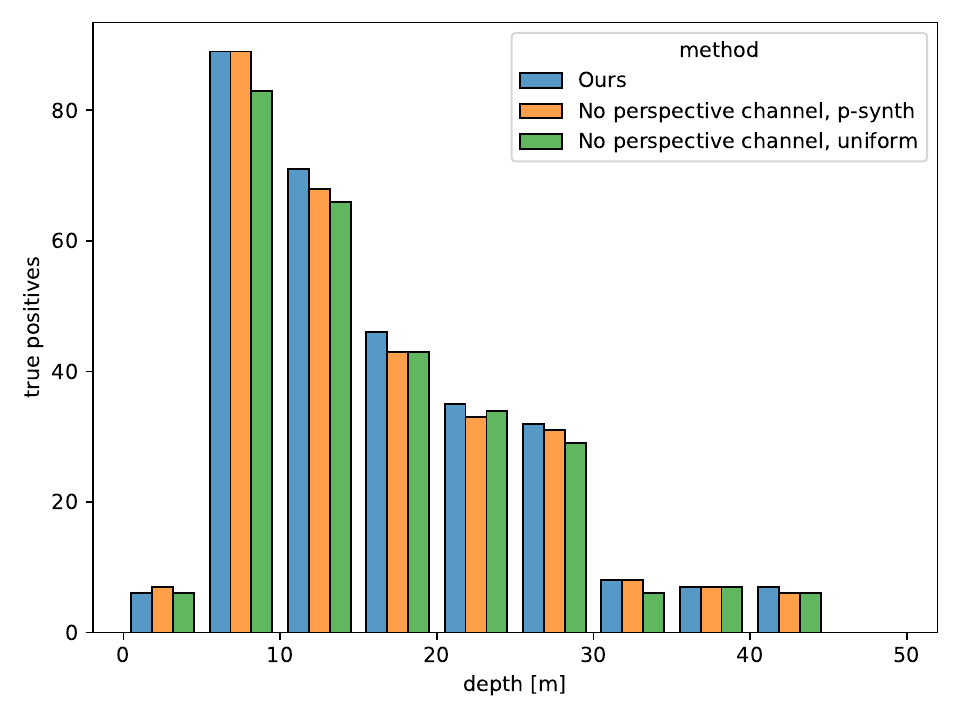}
\caption{\small 
Number of false positives (\textit{left}) and true positives (\textit{right}) as a function of the distance from the camera
for the \textit{Obstacle Track - test} dataset.
Our training set and architecture (Ours) yield much fewer nearby false-positives and slightly more true-positives
than a variant without the perspective map (No perspective channel, p-synth)
or one trained with the uniformly-injected synthetic obstacles (No perspective channel, uniform).
FP and TP are calculated for an IOU threshold of 0.5.
}
\label{fig:instance_distance}
\end{figure*}

This effect is quantified in Fig.~\ref{fig:instance_distance} where we plot the number of false-positives and true-positives as a function of the distance to the camera, estimated as the inverse of the perspective map.
Using our training set and architecture strongly decreases the number of nearby false-positives compared to a system without those contributions, and slightly increases the number of correctly detected obstacles.

\subsubsection{Object size}
\label{sec:ablation_objsize}

In Table~\ref{tab:metrics_ablation_short}-right, we show how the chosen range $\text{Obj}_{[\text{min}, \text{max}]}$ for synthetically injected objects affects detection performance.
To avoid overfitting to the benchmarks, we performed this study using public validation sets, {\it Road Obstacles 21 - Validation} and {\it Lost \& Found - Train}, featuring different obstacles and scenes than the test sets.

As described in Section~\ref{sec:Synthetic-Inject}, the minimum and maximum sizes in meters are multiplied by the local perspective-map value at the site of injection, to determine how big the injected object can be in pixels. We then select at random an object fitting within this pixel range. The results in the table indicate that there is no size range that ideally fits all the circumstances; the small 0.1-0.3 meter range is best for the {\it Lost \& Found - Train} set, while the 0.5-0.9 m range prevails in {\it Road Obstacles 21 - Validation}, presumably due to these ranges matching the typical object sizes in those datasets. We choose the intermediate 0.25-0.55 meter range, 
behaves well in both datasets. One might conclude that including objects of all sizes would be best for generalization, but that would prevent expressing the perspective-size relationship, as shown in the bottom row. Indeed, such a strategy, used in~\cite{Lis19}, yields lower performance than our narrower ranges.

\textbf{Inference Speed.} Our network achieves inference at 12.1 frames of size $1920 \times 1080$ per second on an Nvidia V100 GPU; it can be further sped up by network distillation, quantization or TensorRT.

\section{Conclusion}
\label{sec:conclusion}

We have shown that perspective-aware obstacle injection to generate training data, together with incorporating perspective information in the decoding stages of a network outperforms the state-of-the-art road obstacle detection methods.
Our results indicate that the perspective information can guide the model to reduce false positives for small nearby irregularties while still detecting small and far-away objects.

\clearpage

\bibliographystyle{IEEEtran}
\bibliography{string,vision,learning,biomed,robotics,misc}

\end{document}